\documentclass{llncs}
\usepackage{amsmath}
\usepackage{graphicx,subfigure}
\usepackage{booktabs}
\usepackage{perpage} 
\MakePerPage{footnote} 
\usepackage{multirow}
\DeclareMathOperator*{\argmin}{argmin}
\begin{document}

\title{Real-time image-based instrument classification for laparoscopic surgery}

\author{Sebastian Bodenstedt \inst{1} \and Antonia Ohnemus\inst{1} \and Darko Katic\inst{1} \and
Anna-Laura Wekerle\inst{2} \and Martin Wagner\inst{2} \and Hannes Kenngott\inst{2} \and Beat M{\"u}ller-Stich\inst{2} \and
R{\"u}diger Dillmann\inst{1} \and Stefanie Speidel\inst{1}}
%
%
%
\institute{Institute for Anthropomatics and Robotics, Karlsruhe Institute of Technology,\\
\email{bodenstedt@kit.edu},
\and Department of Abdominal and Transplant Surgery, University of Heidelberg}
\maketitle              

\begin{abstract}
During laparoscopic surgery, context-aware assistance systems aim to alleviate some of the difficulties the surgeon faces.
To ensure that the right information is provided at the right time, the current phase of the intervention has to be known.
Real-time locating and classification the surgical tools currently in use are key components of both an activity-based phase recognition and assistance generation.

In this paper, we present an image-based approach that detects and classifies tools during laparoscopic interventions in real-time. 
First, potential instrument bounding boxes are detected using a pixel-wise random forest segmentation.
Each of these bounding boxes is then classified using a cascade of random forest.
For this, multiple features, such as histograms over hue and saturation, gradients and SURF feature, are extracted from each detected bounding box.

We evaluated our approach on five different videos from two different types of procedures.
We distinguished between the four most common classes of instruments (LigaSure, atraumatic grasper, aspirator, clip applier) and background.
Our method succesfully located up to 86\% of all instruments respectively.
On manually provided bounding boxes, we achieve a instrument type recognition rate of up to 58\% and on automatically detected bounding boxes up to 49\%.

To our knowledge, this is the first approach that allows an image-based classification of surgical tools in a laparoscopic setting in real-time.
\end{abstract}

\section{Introduction}

The goal of a computer-assisted surgery system is to compensate some of the drawbacks typical to laparoscopy by e.g. providing assistance during navigation.
To ascertain what information is currently required, the system has to be context-aware.
This can be accomplished by activity-based phase recognition \cite{lalys2014surgical}\cite{neumuth2011analysis}\cite{KaticIPCAI14}\cite{stauder2014random}.
Commonly, these approaches are based on surgical activities, which generally consist of a tool, an action and an anatomical structure.
One or more given activities can then be used to deduce the current phase of an intervention.

Such a method requires that the surgical tools currently in use have to be located and identified in real-time.
The problem of image-based instrument detection is well known and different approaches can be found in literature \cite{Allan13}\cite{sznitmanMiccai2012a}\cite{voros2007automatic}.
Instrument type classification on the other hand is a lesser known problem.
In \cite{sznitmanMiccai2014} a method for classifying instrument parts, but not explicitly the instrument type, is introduced.
\cite{Speidel2009} introduces a model-based approach for identifying surgical instruments, but does not evaluate the approach on actual surgical videos.

In this paper, we present and evaluate an image-based approach for detecting and classifying instruments during laparoscopic interventions in real-time. 
Our approach relies solely on laparoscopic images, as they are readily available during surgery.
No markers on instruments are required.
Using a random forest classifier, potential instrument bounding boxes are located in each image.
To identify the type of instrument located in each of these bounding boxes, a cascade of two random forest in combination with multiple features, such as hue, saturation, gradients, and SURF, is used.

To our knowledge, our approach is the first that allows a real-time image-based classification of surgical tools in a laparoscopic setting.

For evaluating the approach, we use five video recordings from two different types of laparoscopic surgeries.
\section{Methods}
\begin{figure}[thb]
\centering
 \includegraphics[width=0.8\textwidth]{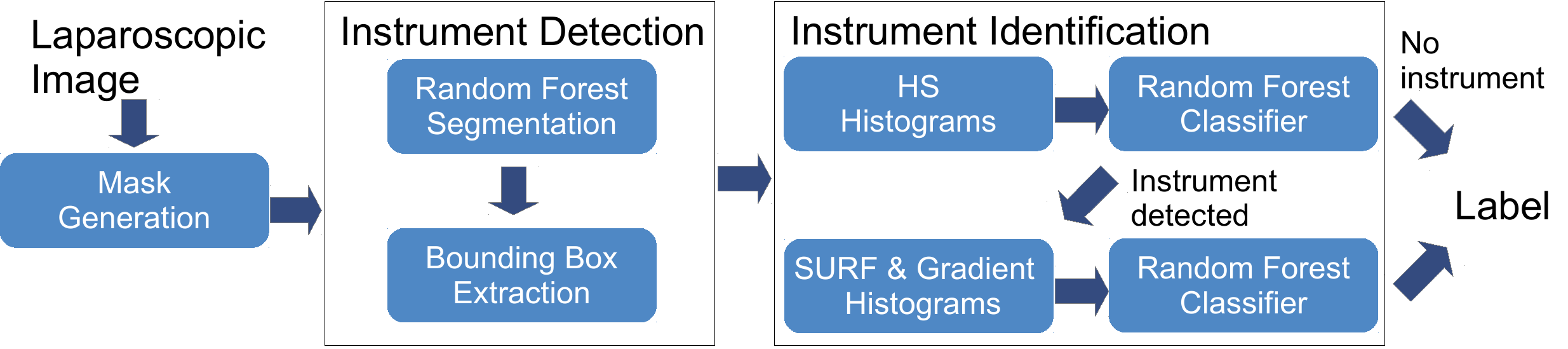}
 \caption{Overview of the detection and identification system. Preprocessed images are first segmented in instrument regions. These regions are then analyzed to ascertain if they really contain an instrument and, if yes, what type.}
 
\label{systemdig}
\end{figure}

An overview of our system for classifying the instruments in a scene can be found in fig. \ref{systemdig}.
First we preprocess the endoscopic image and detect regions that contain surgical instrument candidates.
Each of these regions is examined to ascertain if they actually contain an instrument and, if yes, what type. 
\subsection{Mask Generation}
Since the circular border common to many laparoscopic images has a similar color as most instruments, it can interfere with color features. 
As its size and orientation varies from image to image, we have to detect and remove it automatically.
This is accomplished by applying a threshold filter (threshold = 3) to a grayscale version of the image and then traversing from each corner of the image along the diagonals towards the center until a non-black pixel is found.
The circle is then used to mask the image.
\subsection{Instrument Detection}
\begin{figure}[tb]
\subfigure[]{
\label{ExSeg:Original}
\includegraphics[width=0.31\textwidth]{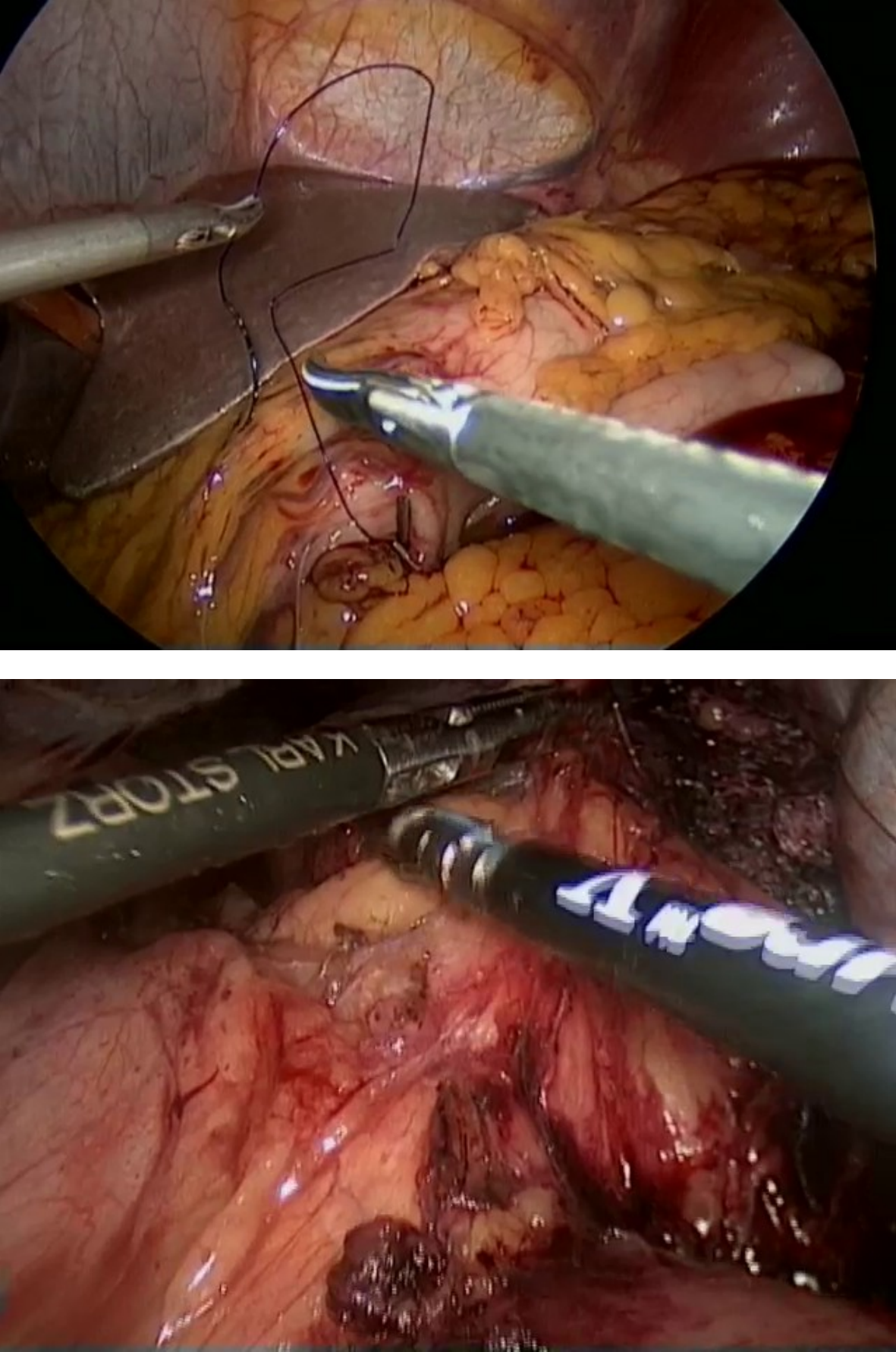}
}
\subfigure[]{
\label{ExSeg:Segmentation}
\includegraphics[width=0.31\textwidth]{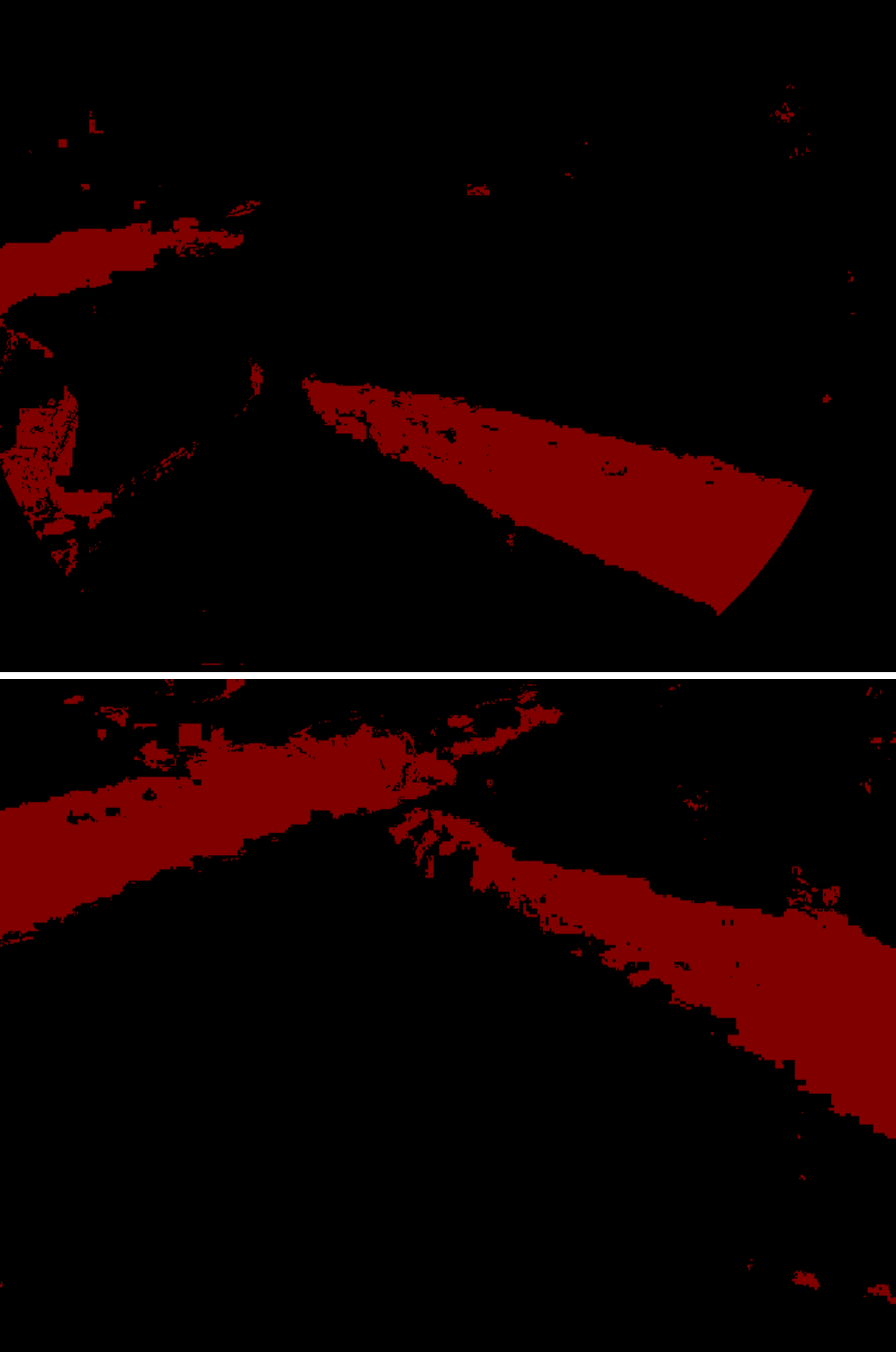}
}
\subfigure[]{
\label{ExSeg:BoundingBox}
\includegraphics[width=0.31\textwidth]{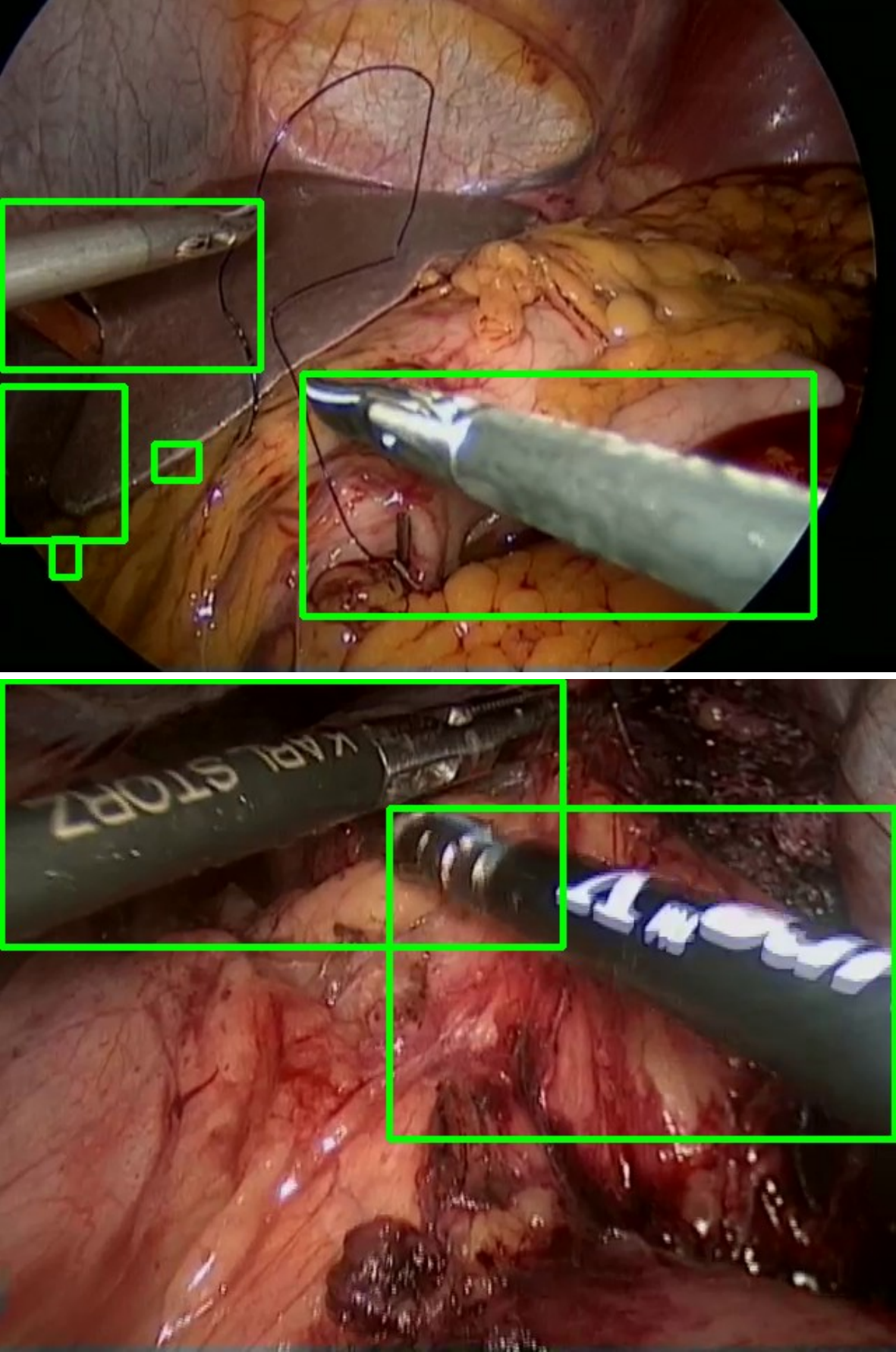}
}

\caption{Two examples of the instrument detection: \subref{ExSeg:Original} Original images. \subref{ExSeg:Segmentation} Output of the random forest. \subref{ExSeg:BoundingBox} bounding boxes after post-processing.}

\label{ExSeg}
\end{figure}

Using a pixel-wise segmentation method, regions of interest that contain surgical instrument candidates are identified.
These regions of interest are given in the form of axis-aligned bounding boxes.
\paragraph{Segmentation}
To localize regions of interest in a surgical scene, a pixel-wise random forest based segmentation method is used \cite{schroff2008object}.
Each pixel in an image is represented through a feature vector consisting of different value from different color spaces, such as RGB, HSV, LAB, opponent and gradient.
To train the random forest, laparoscopic images with previously labeled instruments and background are used as input.  
In order to segment an image in real-time, the random-forest classifier was ported onto GPU.
Examples of the resulting segmentations can be seen in fig. \ref{ExSeg}.
\paragraph{Bounding Box Extraction}
The segmentations are refined with a morphological closing, before locating connected contours.
Gaps on the instrument shaft, e.g. due to blood, have to be closed.
Here we compute the principal direction of each contour by eigen-decomposition of the covariance matrix of all contour points.
Contours that lie in close proximity and share a similar principal direction are fused together.
A bounding box is put around each resulting contour and the five largest candidate boxes are passed on to the classification process.

\subsection{Instrument Identification}
For every bounding box, multiple features are extracted.
Then, using a cascade of two random forests, each box is classified.

As training data, laparoscopic images annotated with labeled bounding boxes around regions of interest are used. 
\paragraph{Image Features}
To describe the content of a bounding box, we use multiple histogram-based features: hue, saturation, gradient and a PCA/SURF\cite{PCASURF} combination.
For each hue and saturation, a uniform 10-bin histogram is used.

\begin{figure}[tb]
\centering
   \includegraphics[width=0.5\textwidth]{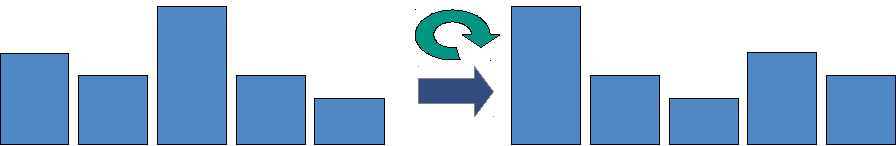}
   \caption{From gradient, an orientation histogram is computed and then aligned according to the highest peak.}
   
   \label{fig:histo}       
\end{figure}

Using gradient, two histograms are constructed.
First, the two sobel operators are applied to the bounding box.
The orientation and magnitude of the gradient for each pixel are calculated from the results and used to construct two 5-bin histograms.
Each orientation is mapped onto the interval $[0^{\circ},180^{\circ}]$ and, to achieve rotation invariance, the bins of the resulting histogram are rotated so that the bin with most entries is at the first position (fig. \ref{fig:histo}).
Further, SURF features \cite{bay2008speeded} are detected in each candidate bounding box and subsequently described with the standard descriptor.
During training, once each frame has been processed, all descriptors are used to perform a principal component analysis (PCA).
Given the $n$ eigenvalues $\lambda_i$ ($\lambda_i \geq \lambda_{i+1}$) of the covariance matrix and a parameter $\alpha$, which determines the amount of variance we want to keep in the data, we compute the number $\hat{m}$ of dimensions to keep:
$$
 \hat{m} = \argmin_m | \alpha\sum_{i=1}^{n} \lambda_i - \sum_{i=1}^{m} \lambda_i|
$$
Each SURF descriptor is rotated with the rotation matrix from the PCA and and the first $\hat{m}$ dimensions are then retained.

All reduced descriptors are then clustered via k-means.
Using the resulting $k$ centers, a histogram with $k$ bins is build by matching each reduced descriptor to the closest center.
\paragraph{Cascading Classification}
To speed-up the classification process, a cascade of two random forests is used.
The first random forest is used to solve the binary problem of determining if a bounding box contains an instrument or not.
Only the hue and saturation histogram are used as feature vector, as they can be computed quickly.
Boxes that are classified as not containing an instrument with a high probability (threshold $\beta$), are discarded.

The remaining boxes are used to train a second random forest, which will determine if the box contains an instrument and distinguish between instrument classes.
Here, in addition to hue and saturation histograms, gradient and PCA-SURF histograms are used.
During classification, these extra histograms are computed only if the bounding box passes the first cascade.
\section{Experiments and Results}
The basis for our evaluation are five laparoscopic videos of two different operation types (two adrenalectomies and three pancreas resections).
To evaluate our approach, we distinguish between the four most commonly occurring classes of instruments: \textit{LigaSure}, \textit{atraumatic grasper}, \textit{aspirator} and \textit{clip applier}.
Furthermore, a fifth class for \textit{no instrument} was added.
Apart from the \textit{atraumatic grasper}, no different instruments of the same type were included in the dataset.
For the \textit{atraumatic grasper}, 3 different instruments were contained in the dataset.
We evaluate the accuracy of the detection and identification methods alone as well as the combination.
Each of the following experiments was repeated ten times and the results were averaged.

Our method was implemented using C++ and the OpenCV library\cite{opencv_library}. 
The evaluation was performed on a workstation PC with an Intel Core i7-5820K, 16GB of RAM and a NVidia Geforce GTX 970.
\paragraph{Instrument Detection}
20 manually selected images from each video (100 overall) were annotated pixel-wise by means of crowd sourcing\cite{maier2013can}.

\paragraph{Instrument Identification}
Two data sets were extracted from the afore mentioned videos.
The first data set consists of manually selected frames, here it was ascertained that at least 10 instances of each class from each video were selected.
The third pancreas resection is an exception, as it did not contain any instances of the aspirator or the clip applier.
For the second data set, 100 frames were taken automatically at a fixed interval from each video.
The data sets were labeled by an expert, as we deemed a medical background necessary for selecting the right tool from a single frame.
Instrument not listed previously was labeled as \textit{unknown}.
The rate of occurrence of each class can be found in table \ref{ev:occ}.
\begin{table}[b]
\caption{The rate of occurrence of each class in the two datasets.}
 
\centering
 \begin{tabular}{l||l|l|l|l|l|l}
  & \textit{No ins.} & \textit{LigaSure} & \textit{A. grasper} & \textit{Aspirator} & \textit{Clip applier}  & \textit{unknown} (Other tools)\\
 \hline
 \hline
Set 1&190&105&145&73&40&48\\
Set 2&311&260&249&55&10&76\\
 \hline
  \hline
 \end{tabular}
  \label{ev:occ}
 \end{table}
 
\subsection{Instrument Detection}
We evaluated the accuracy of the random-forest segmentation method with a leave-one-surgery-out evaluation.
Out of the five available laparoscopic videos, four were used for training and the fifth for testing.

\paragraph{Parameters}
To train the segmentation method, the following parameters were used: As features for each pixel hue, $A$ and $B$ from the LAB color space, $o_1$ and $o_2$ from the opponent color space, and gradient orientation and magnitude are used.
For training the random-forest, 50 trees and a maximum depth of 10 were used.
As a certainty threshold for the background, 60\% was selected.
All parameter values were determined empirically through experiment. 
\paragraph{Results}
The random forest segmentation achieved a precision of 78\%, a recall of 63\% and a DICE coefficient of 0.69 on average.
\subsection{Instrument Detection and Identification}
To assesses the accuracy of the detection and identification framework, we performed two experiments on each the two data sets labeled by expert.
For each run, a leave-one-surgery-out evaluation was performed, training on the data from four operations and evaluating on the fifth.
Set 1 was used for training in every run.

\paragraph{Instrument Identification}
In the first experiment, the manually provided bounding boxes were classified and the results compared to the labels provided by experts.

\paragraph{Detection and Identification}
In the second experiment, the automatically located bounding boxes were classified.
For each detected bounding box, the ground truth bounding box with the highest overlap was located.
If the ratio of the area of the intersection rectangle of the two bounding boxes and the ground truth bounding box exceeded 40\%, we counted the detection as a match and used the annotated label.
When no corresponding bounding box could be located in the labeled data, \textit{no instrument} was used as label.
If a found bounding box was labeled as an \textit{unknown} tool, we counted it as a correct detection, as long as the box was not classified as \textit{no instrument}.
\paragraph{Parameters}
For the random forest, 300 trees and a maximum depth of eight were selected.
Furthermore, $\alpha = 95\%$ and $k = 100$ were used for the PCA-SURF.
During the cascade, $\beta = 60\%$ was used.
All parameter values were determined empirically through experiment. 
\paragraph{Results}

In table \ref{ev:dect}, the number of correctly detected instrument bounding boxes and the number of those correctly identified as a type of instrument (but not necessarily the right type) are listed.
The results of the identification method on the manually and automatically selected bounding boxes can be found in tables \ref{ev:manual} and \ref{ev:auto} respectively, in form of a confusion matrix per set.
We also computed the average percentage of correctly identified classes (table \ref{ev:tpr}).
\begin{table}[tb]
\caption{Total number of instruments, number of instruments detected and number of located instruments correctly identified as any type of instrument in each set averaged over ten runs.}
\centering
 \begin{tabular}{l||c|c|c|c}
&\# Instruments&\# detected&\# classified as instrument\\
 \hline
 \hline
Set 1&411&386 (94\%)&360.5 (88\%)\\
Set 2&650&554.6 (85\%)&517 (80\%)\\
 \hline
 \hline 
 \end{tabular}
  \label{ev:dect}
 \end{table}
 
\begin{table}[tb]
\caption{Confusion matrices illustrating the classification performance on data set 1 \subref{ev:manual:set1} and data set 2 \subref{ev:manual:set2} on manually drawn bounding boxes}
 
\centering
\subfigure[]{
  
\label{ev:manual:set1}
  
\resizebox{0.475\textwidth}{!}{
 \begin{tabular}{|c|l|c|c|c|c|c|}
\cline{3-7}
\multicolumn{2}{c|}{}&\multicolumn{5}{|c|}{Actual class} \\
\cline{3-7}
\multicolumn{2}{c|}{}& \textit{No ins.} & \textit{LigaSure} & \textit{A. grasper} & \textit{Aspirator} & \textit{Clip applier}\\
\hline
\multirow{5}{*}{\rotatebox[origin=c]{90}{Predicted}}&\textit{No ins.}&69\%&3\%&16\%&10\%&1\%\\
\cline{2-7}
&\textit{LigaSure}&1\%&67\%&18\%&8\%&6\%\\
\cline{2-7}
&\textit{A. grasper}&10\%&26\%&44\%&15\%&5\%\\
\cline{2-7}
&\textit{Aspirator}&10\%&13\%&11\%&38\%&28\%\\
\cline{2-7}
&\textit{Clip applier}&4\%&5\%&11\%&10\%&71\%\\
 \hline
 \end{tabular}
 }}
\subfigure[]{
  
\label{ev:manual:set2}
  
\resizebox{0.475\textwidth}{!}{
  \begin{tabular}{|c|l|c|c|c|c|c|}
\cline{3-7}
\multicolumn{2}{c|}{}&\multicolumn{5}{|c|}{Actual class} \\
\cline{3-7}
\multicolumn{2}{c|}{}& \textit{No ins.} & \textit{LigaSure} & \textit{A. grasper} & \textit{Aspirator} & \textit{Clip applier}\\
\hline
\multirow{5}{*}{\rotatebox[origin=c]{90}{Predicted}}&\textit{No ins.}&58\%&4\%&25\%&10\%&3\%\\
\cline{2-7}
&\textit{LigaSure}&2\%&54\%&18\%&15\%&11\%\\
\cline{2-7}
&\textit{A. grasper}&13\%&21\%&43\%&14\%&9\%\\
\cline{2-7}
&\textit{Aspirator}&2\%&14\%&13\%&50\%&21\%\\
\cline{2-7}
&\textit{Clip applier}&18\%&17\%&30\%&0\%&35\%\\
 \hline
 \end{tabular}
 }}

  \label{ev:manual}
 \end{table}
\begin{table}[tb]
\caption{Confusion matrices illustrating the classification performance on data set 1 \subref{ev:auto:set1} and data set 2 \subref{ev:auto:set2} on automatically detected instrument bounding boxes}
  
\centering
\subfigure[]{
\label{ev:auto:set1}
\resizebox{0.475\textwidth}{!}{
 \begin{tabular}{|c|l|c|c|c|c|c|}
\cline{3-7}
\multicolumn{2}{c|}{}&\multicolumn{5}{|c|}{Actual class} \\
\cline{3-7}
\multicolumn{2}{c|}{}& \textit{No ins.} & \textit{LigaSure} & \textit{A. grasper} & \textit{Aspirator} & \textit{Clip applier}\\
\hline
\multirow{5}{*}{\rotatebox[origin=c]{90}{Predicted}}&\textit{No ins.}&50\%&4\%&21\%&25\%&0\%\\
\cline{2-7}
&\textit{LigaSure}&1\%&46\%&16\%&25\%&11\%\\
\cline{2-7}
&\textit{A. grasper}&15\%&20\%&34\%&18\%&12\%\\
\cline{2-7}
&\textit{Aspirator}&6\%&7\%&9\%&58\%&20\%\\
\cline{2-7}
&\textit{Clip applier}&3\%&1\%&7\%&17\%&71\%\\
 \hline
 \end{tabular}
 }}
 \subfigure[]{
\label{ev:auto:set2}
\resizebox{0.475\textwidth}{!}{
  \begin{tabular}{|c|l|c|c|c|c|c|}
\cline{3-7}
\multicolumn{2}{c|}{}&\multicolumn{5}{|c|}{Actual class} \\
\cline{3-7}
\multicolumn{2}{c|}{}& \textit{No ins.} & \textit{LigaSure} & \textit{A. grasper} & \textit{Aspirator} & \textit{Clip applier}\\
\hline
\multirow{5}{*}{\rotatebox[origin=c]{90}{Predicted}}&\textit{No ins.}&51\%&4\%&18\%&26\%&1\%\\
\cline{2-7}
&\textit{LigaSure}&4\%&43\%&17\%&27\%&9\%\\
\cline{2-7}
&\textit{A. grasper}&14\%&17\%&33\%&19\%&17\%\\
\cline{2-7}
&\textit{Aspirator}&1\%&7\%&8\%&62\%&23\%\\
\cline{2-7}
&\textit{Clip applier}&11\%&7\%&33\%&16\%&33\%\\
 \hline
 \end{tabular}
 }
 }
  \label{ev:auto}
 \end{table}
\begin{table}[tb]
\caption{The average percentage of correctly identified tools in each data set.}

\centering
 \begin{tabular}{l||c|c}
&Identification&Detection \& Identification\\
 \hline
 \hline
Set 1&58\%&49\%\\
Set 2&52\%&48\%\\
 \hline
 \hline 
 \end{tabular}
  \label{ev:tpr}
 \end{table}
\subsection{Run-time}
On average, the instrument detection framework required 45ms (22.2Hz) per frame, while the instrument identification took 30ms (32.5Hz).
The times were averaged over 500 runs.
\section{Discussion}
In this paper, we presented, to our knowledge, the first approach for a real-time image-based classification of surgical tools in a laparoscopic setting.
We are currently able to correctly detect 80\% of all instruments in a realistic data set.
Furthermore we are able to correctly determine the type of instrument in 48\% of all cases in the same data set.

\begin{figure}[tb]
\subfigure[]{
\label{Problems:Verdeck}
\includegraphics[width=0.31\textwidth]{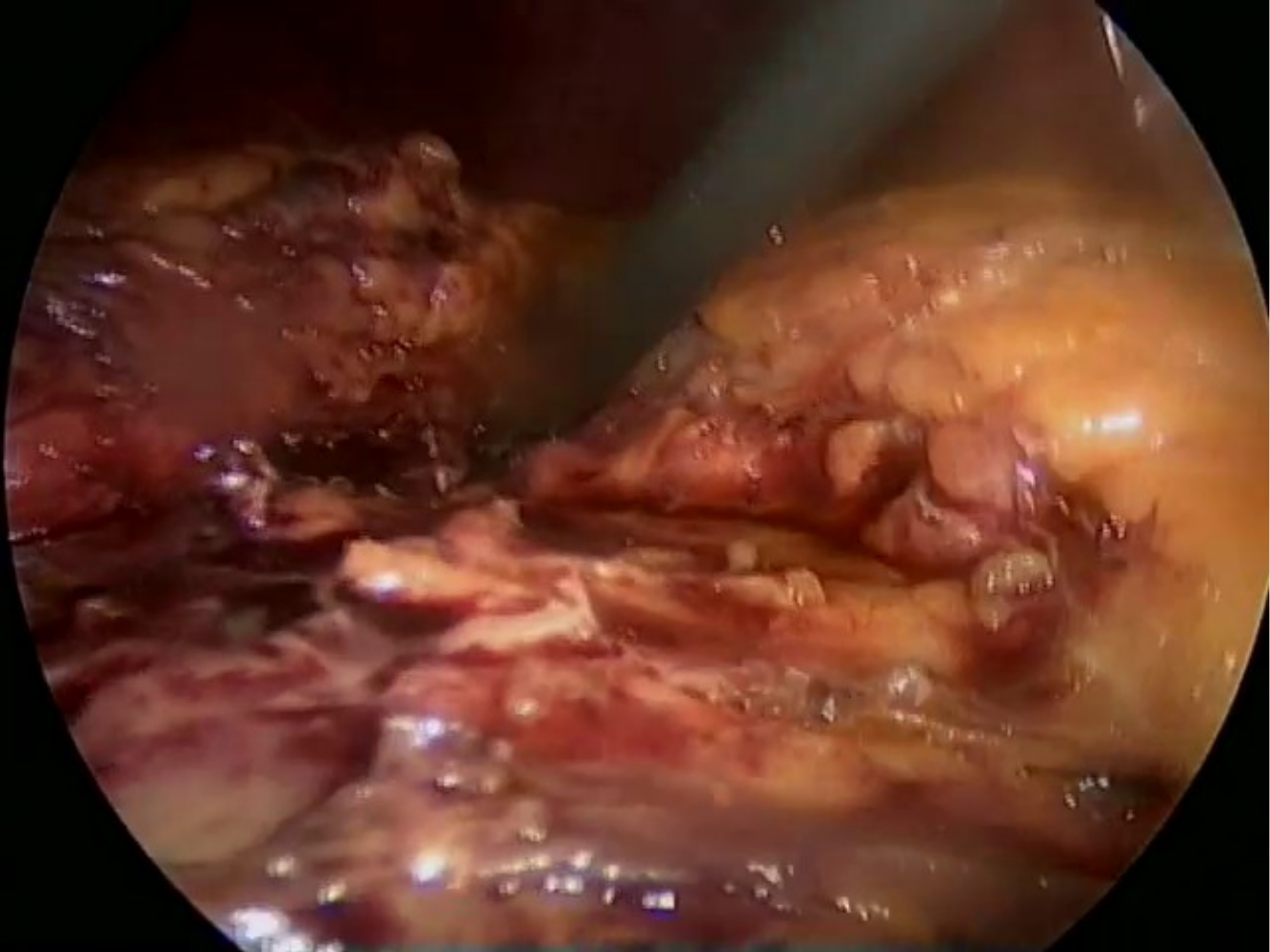}
}
\subfigure[]{
\label{Problems:Overlap}
\includegraphics[width=0.31\textwidth]{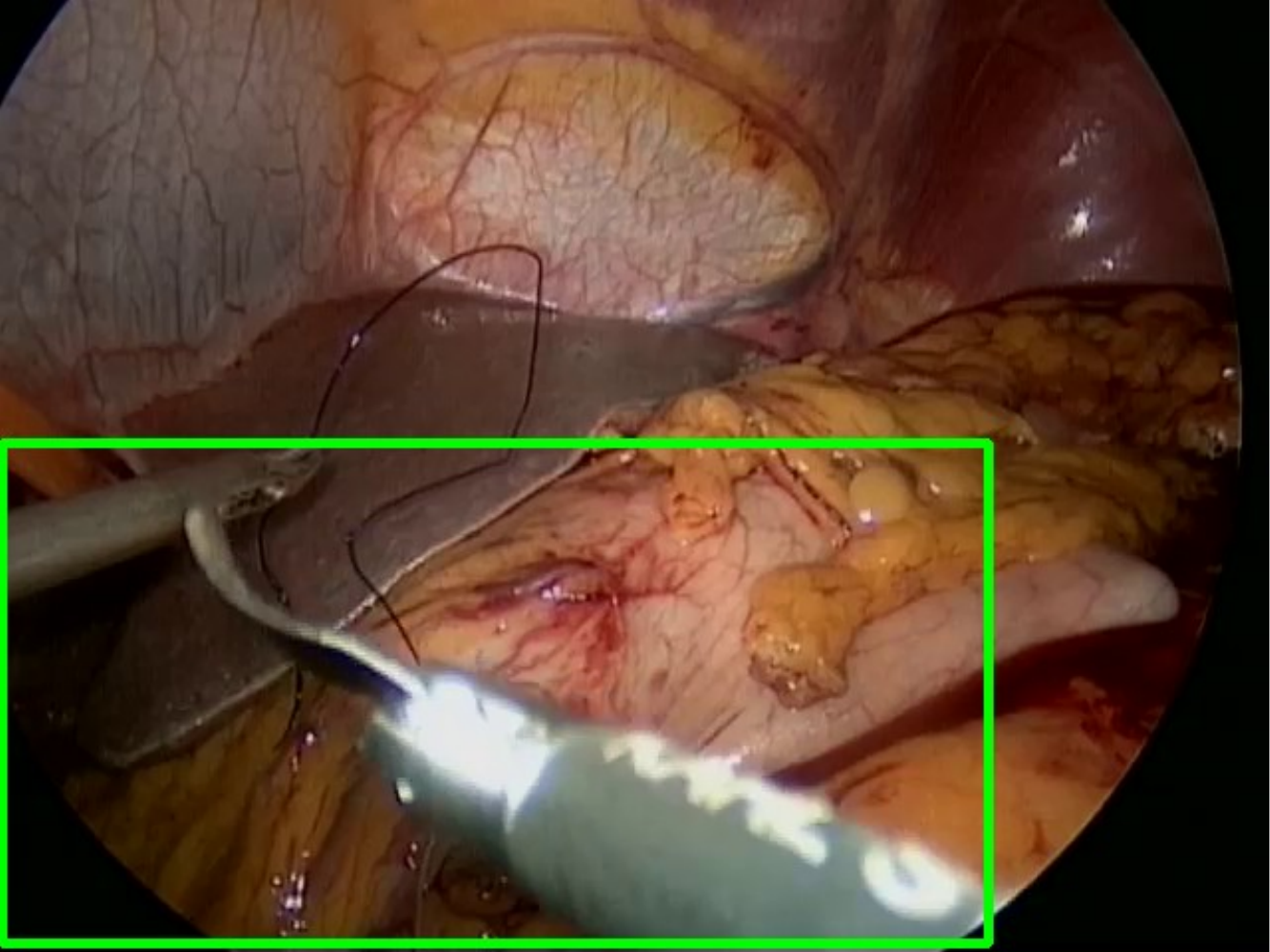}
}
\subfigure[]{
\label{Problems:Motion}
\includegraphics[width=0.31\textwidth]{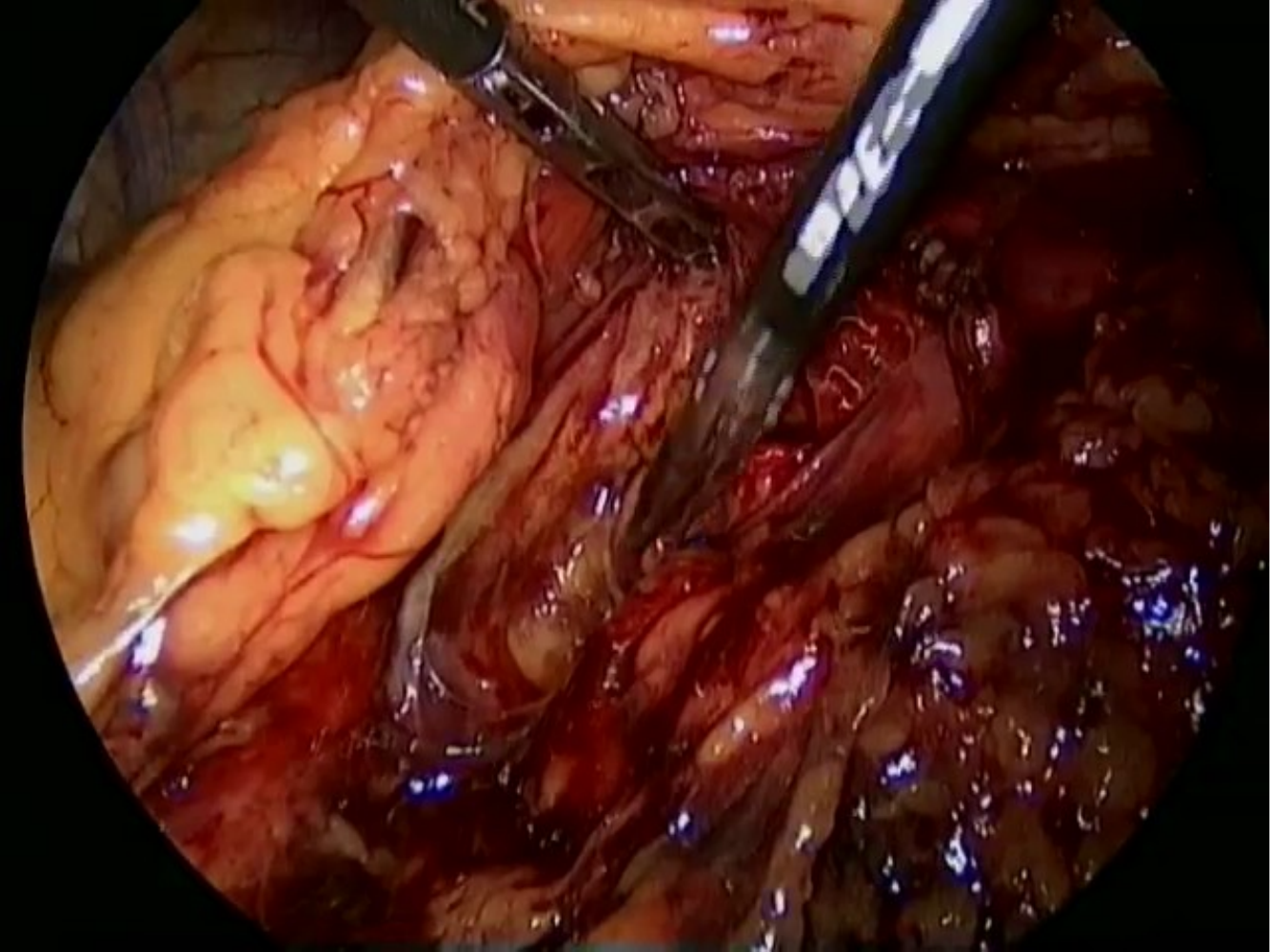}
}

\caption{Potential error sources: \subref{Problems:Verdeck} Instrument tip not visible, \subref{Problems:Overlap} two overlapping instruments detected as one and \subref{Problems:Motion} motion blur.}

\label{Problems}
\end{figure}

Some of the major problems that occur, especially on realistic data, can be seen in fig. \ref{Problems}.
If part of an instrument, e.g. the tip (fig. \ref{Problems:Verdeck}), is occluded by tissue or blood, ambiguities are possible, since different types of instrument share a similar shaft and can therefore only be reliably distinguished by the tip.
Motion blur (fig. \ref{Problems:Motion}) can also cause ambiguities.
If two instruments overlap (fig. \ref{Problems:Overlap}) they can be detected as one instrument.
Further error sources are differences in illumination and white balance, which can vary between different operations, especially if a different optic is used.
The most common confusions were between \textit{LigaSure}, \textit{atraumatic grasper} and \textit{aspirator} as they share a similar formed shaft, which, under different lighting conditions, can be difficult to distinguished.
Also, when only a small portion of the instrument could be found, either due to occlusions or due to it just entering the field of view, a confusion with the \textit{no instrument} class was frequent.

As a solution to the majority of these problems, future research will focus on incorporating tracking methods to propagate succesfully detected and identified instruments over time.
This would mitigate the effect of occluded instrument tips, assuming the tip was visible in previous frames.
A study to determine if incorporating a larger variety in laparoscopic videos during training would increase the detection rate, is planned.

\section{Acknowledgements}
\label{acknowledgments}
The present research was conducted within the setting of the SFB/Transregio 125 ``Cognition-Guided Surgery'' funded by the German Research Foundation. It is furthermore sponsored by the European Social Fund of the State Baden-Wuerttemberg.
\bibliography{paper}

\begin{thebibliography}{10}
\providecommand{\url}[1]{\texttt{#1}}
\providecommand{\urlprefix}{URL }

\bibitem{Allan13}
Allan, M., Ourselin, S., Thompson, S., Hawkes, D., Kelly, J., Stoyanov, D.:
  Toward detection and localization of instruments in minimally invasive
  surgery. Biomedical Engineering, IEEE Transactions on  60(4),  1050--1058
  (April 2013)

\bibitem{bay2008speeded}
Bay, H., Ess, A., Tuytelaars, T., Van~Gool, L.: Speeded-up robust features.
  Computer vision and image understanding  (2008)

\bibitem{opencv_library}
Bradski, G.: Dr. Dobb's Journal of Software Tools

\bibitem{PCASURF}
Gonzalez~Valenzuela, R., Robson~Schwartz, W., Pedrini, H.: Dimensionality
  reduction through pca over sift and surf descriptors. In: Proc. of IEEE CIS
  (2012)

\bibitem{KaticIPCAI14}
Katic, D., Wekerle, A.L., G\"{a}rtner, F., Kenngott, H., M\"{u}ller-Stich, B.,
  Dillmann, R., Speidel, S.: Knowledge-driven formalization of laparoscopic
  surgeries for rule-based intraoperative context-aware assistance. In:
  Stoyanov, D., Collins, D., Sakuma, I., Abolmaesumi, P., Jannin, P. (eds.)
  Information Processing in Computer-Assisted Interventions, Lecture Notes in
  Computer Science, vol. 8498, pp. 158--167. Springer International Publishing
  (2014)

\bibitem{lalys2014surgical}
Lalys, F., Jannin, P.: Surgical process modelling: a review. IJCARS  (2014)

\bibitem{maier2013can}
Maier-Hein, L., Mersmann, S., Kondermann, D., Bodenstedt, S., Sanchez, A.,
  Stock, C., Kenngott, H., Eisenmann, M., Speidel, S.: Can masses of
  non-experts train highly accurate image classifiers? In: Golland, P., Hata,
  N., Barillot, C., Hornegger, J., Howe, R. (eds.) Medical Image Computing and
  Computer-Assisted Intervention – MICCAI 2014. Lecture Notes in Computer
  Science, vol. 8674, pp. 438--445. Springer International Publishing (2014)

\bibitem{neumuth2011analysis}
Neumuth, T., Jannin, P., Schlomberg, J., Meixensberger, J., Wiedemann, P.,
  Burgert, O.: Analysis of surgical intervention populations using generic
  surgical process models. International Journal of Computer Assisted Radiology
  and Surgery  6(1),  59--71 (2011),
  \url{http://dx.doi.org/10.1007/s11548-010-0475-y}

\bibitem{schroff2008object}
Schroff, F., Criminisi, A., Zisserman, A.: Object class segmentation using
  random forests. In: Proc. of BMVC (2008)

\bibitem{Speidel2009}
Speidel, S., Benzko, J., Krappe, S., Sudra, G., Azad, P., M{\"u}ller-Stich,
  B.P., Gutt, C., Dillmann, R.: Automatic classification of minimally invasive
  instruments based on endoscopic image sequences. In: SPIE Medical Imaging
  (2009)

\bibitem{stauder2014random}
Stauder, R., Okur, A., Peter, L., Schneider, A., Kranzfelder, M., Feussner, H.,
  Navab, N.: Random forests for phase detection in surgical workflow analysis.
  In: Stoyanov, D., Collins, D., Sakuma, I., Abolmaesumi, P., Jannin, P. (eds.)
  Information Processing in Computer-Assisted Interventions, Lecture Notes in
  Computer Science, vol. 8498, pp. 148--157. Springer International Publishing
  (2014)

\bibitem{sznitmanMiccai2012a}
Sznitman, R., Ali, K., Richa, R., Taylor, R., Hager, G., Fua, P.: Data-driven
  visual tracking in retinal microsurgery. In: Ayache, N., Delingette, H.,
  Golland, P., Mori, K. (eds.) Medical Image Computing and Computer-Assisted
  Intervention – MICCAI 2012, Lecture Notes in Computer Science, vol. 7511,
  pp. 568--575. Springer Berlin Heidelberg (2012)

\bibitem{sznitmanMiccai2014}
Sznitman, R., Becker, C., Fua, P.: Fast part-based classification for
  instrument detection in minimally invasive surgery. In: Proc. of MICCAI
  (2014)

\bibitem{voros2007automatic}
Voros, S., Long, J.A., Cinquin, P.: Automatic detection of instruments in
  laparoscopic images: A first step towards high-level command of robotic
  endoscopic holders. IJRR  (2007)

\end{thebibliography}
\bibliographystyle{splncs03}

\end{document}